%% file: inavit.tex
\documentclass{article}

\usepackage[verbose=true,letterpaper]{geometry}
\AtBeginDocument{
  \newgeometry{
    textheight=9in,
    textwidth=6in,
    top=1in,
    headheight=12pt,
    headsep=25pt,
    footskip=30pt
  }  
}

\usepackage[utf8]{inputenc} 
\usepackage[T1]{fontenc}    
\usepackage{url}            
\usepackage{booktabs}       
\usepackage{amsfonts}       
\usepackage{nicefrac}       
\usepackage{microtype}      
\usepackage{xcolor}         
\usepackage{graphicx}
\usepackage{multirow}
\usepackage{amsmath, calc}
\usepackage{dirtytalk}
\usepackage{algorithm2e}
\usepackage{algpseudocode}
\usepackage{authblk}
\usepackage{xr}
\usepackage{caption}
\usepackage{subcaption}

\usepackage[capitalize]{cleveref}
\crefname{section}{Sec.}{Secs.}
\crefname{section}{Section}{Sections}
\crefname{table}{Table}{Tables}
\crefname{table}{Tab.}{Tabs.}

\title{Interaction Region Visual Transformer for Egocentric Action Anticipation}

\author[1]{Debaditya Roy}
\author[1]{Ramanathan Rajendiran}
\author[1,2]{Basura Fernando}
\affil[1]{Institute of High Performance Computing (IHPC), Agency for Science, Technology and Research (A*STAR), 1 Fusionopolis Way, \#16-16 Connexis, Singapore 138632, Republic of Singapore.}
\affil[2]{Centre for Frontier AI Research (CFAR), Agency for Science, Technology and Research (A*STAR), 1 Fusionopolis Way, \#16-16 Connexis, Singapore 138632, Republic of Singapore}
\date{} 

\begin{document}
\maketitle

\begin{abstract}
 Human-object interaction is one of the most important visual cues and we propose a novel way to represent human-object interactions for egocentric action anticipation. We propose a novel Transformer variant to model interactions by computing the change in the appearance of objects and human hands due to the execution of the actions and use those changes to refine the video representation. Specifically, we model interactions between hands and objects using Spatial Cross-Attention (SCA) and further infuse contextual information using Trajectory Cross-Attention to obtain environment-refined interaction tokens. Using these tokens, we construct an interaction-centric video representation for action anticipation. We term our model InAViT which achieves state-of-the-art action anticipation performance on large-scale egocentric datasets EPICKTICHENS100 (EK100) and EGTEA Gaze+. InAViT outperforms other visual transformer-based methods including object-centric video representation. On the EK100 evaluation server, InAViT is the top-performing method on the public leaderboard (at the time of submission) where it outperforms the second-best model by 3.3\% on mean-top5 recall.
\end{abstract}

\section{Introduction}
\input{new_intro.tex}

\section{Related Work}
\input{related_work.tex}
\section{Preliminaries}
\input{model_1}
\section{Our method}\label{sec:modelint}
\input{model_2}

\input{model_3}

\section{Experiments and Results}
\input{exp_new}

\section{Discussions and Conclusion}
\input{conclusion}

\section*{Acknowledgment}
This research/project is supported in part by the National Research Foundation, Singapore under its AI Singapore Program (AISG Award Number: AISG-RP-2019-010) and
by the National Research Foundation Singapore and DSO National Laboratories under the AI Singapore Programme (AISG Award No: AISG2-RP-2020-016).


\input{sup}
{\small
\bibliographystyle{ieee_fullname}
\bibliography{inavit}
}

\end{document}

%% file: new_intro.tex
In egocentric action anticipation, the model needs to predict the immediate next human action that is going to happen, usually 1 second into the future~\cite{furnari2020rolling}.
Action anticipation is a challenging task due to various reasons such as the \textit{uncertainty in future actions}, \textit{diversity of execution of actions}, and the \textit{complexity of human-object interactions} presented when executing those actions.
A way to reduce the uncertainty in predicting the next action is to develop models that may be able to infer information about probable future objects and interactions that will be used in the future.
As the observed and next actions are causally related, it is highly likely that information about observed interactions may possess possible cues about future actions.

Some of the existing works use hand-object interactions to exploit those cues for action anticipation~\cite{roy2021action,liu2020forecasting}, yet do not exploit the visual changes in human and object appearance caused by the execution of actions.
We hypothesize that modeling of change in the appearance of regions containing objects and hands may reveal vital information about the probable execution of future actions.
The work in \cite{liu2020forecasting} focuses on predicting manually annotated interaction hotspots without accounting for the specific objects associated with the interaction. 
However, one of the biggest challenges in incorporating interactions from egocentric videos is to identify and capture the most relevant image patches among the entire hand-object interaction region that participates in the ongoing action.
We develop an approach that attends to the most informative patches among the entire interaction region in every observed frame and make use of those modeled changes in appearances to predict future actions.

In particular, we propose to model the interaction region containing the hands and the interacted objects. 
We model the interaction regions as how hand and object appearance change due to the execution of the action and use those changes in appearance.
Furthermore, we infuse the interaction regions with contextual cues from the background of the current action to obtain additional visual information about the area in which the interactions are performed.
Finally, we propose an effective way to incorporate context-infused hand-object interaction regions to a video Transformer to create a richer interaction-centric video representation.
Specifically, we use effective MotionFormer~\cite{patrick2021keeping} as it aggregates important dynamic information along implicitly determined motion paths.
This simple yet effective idea improves the action anticipation performance outperforming many dedicated video Transformer models.

Leveraging on the MotionFormer~\cite{patrick2021keeping}, we design a spatio-temporal visual transformer denoted as human-object Interaction visual transformer (InAViT) for action anticipation that refines image patches from the interaction regions in every frame which we term as \textit{interaction tokens}. 
The interaction tokens are obtained from the refined object and human tokens and the refinement is influenced by objects, hands, and the anticipated action.
We incorporate the visual context of the interaction's surroundings into the interaction tokens by proposing a trajectory cross-attention mechanism based on trajectory attention~\cite{patrick2021keeping}.
Finally, we infuse interaction tokens into the observed video to build an interaction-centric video representation for effective action anticipation.
InAViT provides a way to extract visual changes in interaction regions across observed frames using trajectory cross-attention while in ~\cite{roy2021action}, human and object visual features are simply concatenated frame-wise to represent interactions.
 
Human-object interactions in actions are expressed as relations in spatio-temporal graphs in~\cite{jain2016structural, wang2018videos, materzynska2020something, teng2021target, ou2022object}. 
We are inspired by these approaches, especially the progression of human-object relationships for action recognition tasks~\cite{ou2022object}.
We model interaction regions explicitly as spatio-temporal visual changes in hands and objects rather than implicitly as edges between hands and objects~\cite{ou2022object}. 
Furthermore, authors in ~\cite{herzig2022object, zhang2022object} make use of only object dynamics,
and do not treat human hands as a separate entity~\cite{materzynska2020something,ou2022object}.
On the other hand, our interaction-centric video representation recognizes hands as a separate entity from objects that affect visual change on objects and vice-versa.
The change in the hand's visual appearance when interacting with objects also gives a clue about the observed action.
Hence, modeling interaction regions by capturing visual changes in both hands and objects is better than modeling the change in objects alone \cite{herzig2022object}. 
Therefore, we postulate that interaction-centric representations are better suited for action anticipation than object-centric approaches.

In summary, our contributions are twofold:
(1) We propose a novel human-object interaction module that computes appearance changes in objects and hands due to the execution of the action and models these changes in appearances to refine the video representations of Video Transformer models for effective action anticipation.
(2) On the EK100 evaluation server, our model \textbf{InAViT is top of the public leaderboard} and outperforms the second-best model by $3.3\%$ on mean-top5 recall. Leveraging on MotionFormer, we also obtain massive improvement in EGTEA Gaze+ dataset.

%% file: related_work.tex
Anticipating human actions has gained interest in the research community with large datasets~\cite{grauman2022ego4d,min2021integrating,damen2022rescaling} and innovative approaches~\cite{furnari2020rolling,sener2020temporal,girdhar2021anticipative,wu2022memvit,fernando2021,gu2021transaction,wu2021Anticipate, qi2021self,zatsarynna2021multi, ke2019time,gammulle2019predicting,nmns}. 
In~\cite{furnari2020rolling}, rolling and unrolling LSTM (RU-LSTM) is proposed to predict the next action. 
~\cite{sener2020temporal} increases RU-LSTM's temporal context using non-local blocks to combine local and global temporal context. 
In~\cite{girdhar2021anticipative}, spatio-temporal transformers called Anticipative Video Transformers (AVT) are proposed for action anticipation.
In~\cite{wu2022memvit}, MemViT extends AVT for long-range sequences by memory caching multiple smaller temporal sequences.
RAFTformer \cite{girase2023latency} proposes a real-time action anticipation transformer using learnable anticipation tokens to capture global context trained using masking-based self-supervision.
In~\cite{gong2022future,nawhal2022rethinking}, transformers are used for long-term anticipation (LTA).
In~\cite{dessalene2023therbligs}, motion primitives called therbligs are introduced for decomposing actions which are then used for action anticipation.
~\cite{Roy_2022_WACV, roy2022predicting} focus on modeling a goal representation for next action anticipation while \cite{mascaro2023intention} focuses on discovering intentions for LTA.
\cite{zhong2023anticipative} uses audio information to augment video representation for next-action anticipation and \cite{mittallearning} uses only audio for LTA.
Other approaches consider past and future correlation using Jaccard vector similarity~\cite{fernando2021}, self-regulated learning~\cite{qi2021self}, transitional model~\cite{leveraging}, and counterfactual reasoning~\cite{zhangif21}.

There are a variety of approaches for Human-Object Interactions (HOIs) modeling in images \cite{chao2018learning, gao2018ican, li2019transferable, gao2020drg, zhang2021mining, chen2021reformulating, kim2020uniondet, kim2021hotr,liao2020ppdm} and videos ~\cite{liu2020forecasting,li2021weakly,liu2022joint,ji2021detecting, faure2023holistic}.
In ~\cite{li2021weakly}, human-object interaction regions are detected in videos using verb-object queries describing the action. 
Observed action labels are not available during testing in action anticipation and hence, we cannot predict the interaction region.
In~\cite{ji2021detecting}, relationships between humans and objects are modeled and verified using relationship labels.
Relationship labels are not available for the observed video and so our model attends to all interactions and discovers their importance in predicting the next action.

Another related area is interaction hotspot prediction where future hand-trajectory and interaction spots need to be estimated. 
Interaction prediction approaches~\cite{liu2020forecasting, liu2022joint} learn future hand motion distribution conditioned on the video representation using an encoder (LSTM~\cite{liu2020forecasting} or Transformer~\cite{liu2022joint}).
Object interaction anticipation \cite{ragusa2023stillfast} requires predicting the next active object bounding box along with the next action.
However, these approaches require explicit annotations of hand trajectories in future, object trajectories in the observed frames and location of interaction spots.
Our interaction modeling approach obviates the need for hand and object trajectory annotations that are difficult to obtain in videos as shown in ~\cite{fouhey2018lifestyle}. 

%% file: model_1.tex

\subsection{Basic video representation}
We extract a set of 3-D cuboids or video "tokens" from a video as existing video transformers~\cite{arnab2021vivit,patrick2021keeping,bertasius2021space}.
The set of all the video tokens from a single fixed length video-clip is denoted by $X \in \mathbb{R}^{THW \times d}$ wherein each of the $THW$ cuboids is linearly projected to a $d$-dimensional vector.
Here, $T$ is the number of frames in the fixed-length video clip and $H, W$ is the number of vertical and horizontal patches respectively.
Let $\mathbf{x}_{st} \in \mathbb{R}^d$ denote a video token from the set $X$ at spatial location $s \in \{1, \cdots, H \times W\}$ and temporal location $t \in \{1, \cdots, T\}$.
Similar to~\cite{patrick2021keeping}, we add separate learnable positional encoding for spatial and temporal dimension for each video token denoted as $\mathbf{e}_s^s \in \mathbb{R}^d$  and $\mathbf{e}_t^t \in \mathbb{R}^d$, respectively.
The resultant video token after spatial and temporal embedding is given as $\mathbf{x}_{st} = \mathbf{x}_{st} + \mathbf{e}_s^s + \mathbf{e}_t^t$.
A classification token $\mathbf{x}_{cls}$ is appended for anticipating the next action from $X$ resulting in $THW + 1$ tokens in $\mathbb{R}^d$.
We exclude the classification token hereafter for clarity.

\subsection{Obtaining hand and object tokens}~\label{sec:hotokens}
We obtain hand and object representations from video tokens $X$.
We obtain object and hand bounding-boxes using Faster R-CNN~\cite{ren2015faster}.
In every frame, we use one bounding box for the hand and $N$ bounding boxes for objects closest to the hand.
SORT algorithm is used over the detections to obtain sequences of detections where each sequence represents a hand or an object~\cite{ou2022object}.
Now, given the video tokens corresponding to a frame $X_t$ and the bounding box of the hand $B_{h,t}$, we obtain a hand token $\mathbf{h}_t \in \mathbb{R}^d$. We make use of RoIAlign~\cite{he2017mask} layer on $X_t$ to obtain hand region crops similar to~\cite{herzig2022object}.
We then use MLP and max-pooling to obtain the final hand representation or the hand token.
We apply this to every frame to obtain $T$ hand tokens denoted by $\mathcal{H} \in \mathbb{R}^{T\times d}$ where $ \mathcal{H} = [\mathbf{h}_1, \cdots, \mathbf{h}_T]$.
Similarly, for each of object $i$, we obtain a  $T$ object tokens $\mathbf{o}_{i,1}, \cdots, \mathbf{o}_{i,T}$ and we denote it as $\mathcal{O}_i \in \mathbb{R}^{T\times d}$.
In total, for the $N$ objects, we end up with $\mathcal{O} \in \mathbb{R}^{T\times N \times d}$ where $ \mathcal{O}= [\mathcal{O}_1, \cdots, {\mathcal{O}}_N]$.

%% file: model_2.tex
\subsection{Overview of our method}
We hypothesize that in egocentric action anticipation, hands and objects play a key role in anticipating actions. 
Hands and objects change the appearance of other objects causing visible state changes, such as \textit{human cutting tomato with knife} or \textit{human emptying a pan using spatula}. 
Change in the state of the objects reveals cues about the possible next action.
We capture these changes using newly designed \emph{interaction tokens} by refining the original hand and object representations (tokens) with respect to each other.

As the objects affect the appearance of hand regions, we refine the hand tokens using all object tokens in the frame using \cref{hchange}.
Similarly, as hand and objects affect the appearance of other objects when executing the action, we model this by refining every object token using hand tokens and other object tokens in the frame using \cref{ochange}.
\begin{align}
    &\mathcal{\tilde{H}} = \phi_\mathcal{H}(\mathcal{H}|\mathcal{O}) \label{hchange}\\    
    &\mathcal{\tilde{O}}_{i} = \phi_\mathcal{O}(\mathcal{O}_{i}|\mathcal{H}, \mathcal{O}_j) \text{  } \forall j \ne i, i \in {1, \cdots,N} \label{ochange},
\end{align}
Here $\phi_\mathcal{H}$ and $\phi_\mathcal{O}$ are attention-based functions that will be discussed in detail in \cref{sec:sca,sec:sot,sec:ub}.
Refined object tokens for all objects are denoted as $\mathcal{\tilde{O}} = [\mathcal{\tilde{O}}_1, \cdots, \mathcal{\tilde{O}}_N]$. 
Together, the refined hand and object tokens constitute the interaction tokens $I = [\mathcal{\tilde{H}}, \mathcal{\tilde{O}}]$.

\begin{figure}[t]
    \centering
    \includegraphics[width=0.8\linewidth]{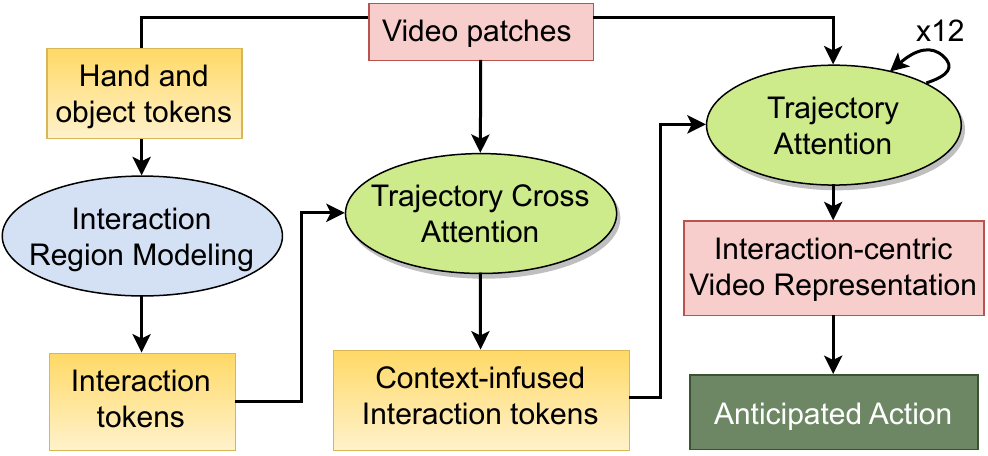}
     \caption{Block diagram of our interaction region modeling based action anticipation (InAVIT). Ellipses represent processing mechanisms. Trajectory attention $\times12$ refers to MotionFormer.}
    \label{fig:fullblock}
\end{figure}

We also hypothesize that the context or the background of the current action may provide useful information when predicting the next action.
For example, \textit{picking a tomato} next to the cutting board (context) informs that the next probable action is \textit{cut tomato}.
Therefore, we enrich the interaction tokens ($I$) using the information coming from the context and vice-versa.
We use Trajectory Cross Attention function ($\phi_I$) inspired by~\cite{patrick2021keeping} to obtain the context-infused interaction tokens $\tilde{I}$
\begin{equation}
\tilde{I} = \phi_I(I| X).   
\label{eq.tca.overview}
\end{equation}
The final video representation is interaction-centric where we first concatenate context-infused interaction tokens to the video tokens. 
Then, we assimilate the interaction regions into the video using self-attention ($\phi_X$).
We choose the refined tokens corresponding to the original video tokens$X$ as the interaction-centric video representation $X_I$
\begin{equation}
    X_I = \phi_X( [\tilde{I}, X]). \label{vchange}
\end{equation}
We predict the next action using the interaction-centric video representation with multiple layers of Trajectory Attention as in MotionFormer~\cite{patrick2021keeping}
\begin{equation}
    \mathbf{a}_{next} = \phi(X_I).
\end{equation}
Next action anticipation is defined as observing $1, \cdots, T$ frames and predicting the action that happens after a gap of $T_a$ seconds.
It is important to note that a new action starts after $T_a$ seconds that is not seen in the observed frames.
Our overall approach is shown in \cref{fig:fullblock}. 
\begin{figure}[t]
    \centering
    \includegraphics[width=\linewidth]{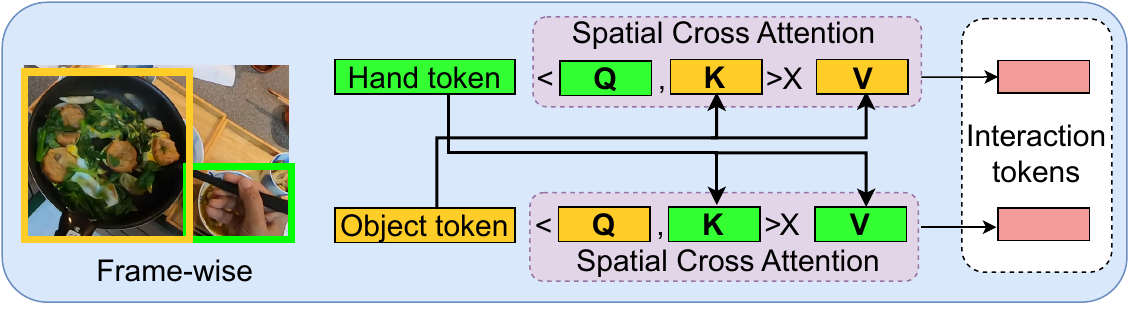}
    \caption{Modeling interaction region tokens using Spatial Cross Attention. In every frame, hand tokens act as query and object tokens as key and value to compute refined hand tokens. Refined object tokens are computed with object token as query, and hand and other object tokens as key and values (not shown here to avoid clutter). Interaction tokens consist of refined hand and object tokens.}
    \label{fig:modelint_s}
\end{figure}

\subsection{Human-Object interaction region modeling}
Now we discuss how we implement ~\cref{hchange} and ~\cref{ochange}.
We model spatiotemporal interaction regions between hands and objects in three ways encapsulating different types of interaction information to obtain \emph{interaction tokens} $I = [\mathcal{\tilde{H}}, \mathcal{\tilde{O}}]$. These three types of interaction modeling will be evaluated in the experiments.
Reader may refer to the Appendix A for the common definition of cross-attention and self-attention.
Next, we present three ways to obtain interaction tokens.
\subsubsection{SCA: Spatial cross-attention}
\label{sec:sca}
We implement ~\cref{hchange} and ~\cref{ochange} using spatial cross-attention.
In every observed frame, there is a hand token and multiple object tokens.
We model the change in hands by cross-attention~\cite{vaswani2017attention} as shown in \cref{fig:modelint_s}.
We use the hand token  $\mathbf{h}_{t}$ as the query and compute the attention with respect to every object token in the same frame $\mathbf{o}_{1,t},\cdots, \mathbf{o}_{N,t}$.
The query, key, and value are denoted as $\mathbf{q}_{h,t} = \mathbf{h}_{t} \mathbf{W}_{q},~~\mathbf{k}_{i,t} =  \mathbf{o}_{i,t} \mathbf{W}_{k},~~\text{ and } \mathbf{v}_{i,t} = \mathbf{o}_{i,t} \mathbf{W}_{v}$, respectively.
We use cross-attention to get the refined hand tokens, $\tilde{\mathbf{h}}_t$.
Cross attention implicitly seeks the object token that has the most impact on the hand token by pooling all the object tokens in the frame and weighing each by its probability.

Similarly, we use cross-attention to refine each object token using the hand token and other object tokens in the frame. 
Every object token $\mathbf{o}_{i,t}$ acts as a query, and human and other object tokens act as keys and values. 
Let $\mathbf{z}_{t}$ represent either hand or other object tokens in the frame $t$. 
There are $N$ such tokens in each frame for each object query $\mathbf{o}_{i,t}$.
The query, key, and values are obtained as
$\mathbf{q}_{i,t} = \mathbf{o}_{i,t} \mathbf{W}_{q},~~\mathbf{k}_{j,t} =  \mathbf{z}_{t} \mathbf{W}_{k}, \text{ and } ~\mathbf{v}_{j,t} = \mathbf{z}_{t} \mathbf{W}_{v}$, respectively.
The refined object token $\tilde{\mathbf{o}}_{i,t}$ are obtained using cross-attention. 
We call the refined hand and object tokens as interaction tokens $I_t = [\tilde{\mathbf{h}}_t, \tilde{\mathbf{o}}_{1,t},\cdots,\tilde{\mathbf{o}}_{N,t}]$.
We perform spatial cross-attention (SCA) over every frame to obtain all the interaction tokens $I \in \mathbb{R}^{T \times (N+1)\times d}$.

\subsubsection{SOT: Self-attention of hand/object over time}~\label{sec:sot} 
\begin{figure}
    \centering
    \includegraphics[width=\linewidth]{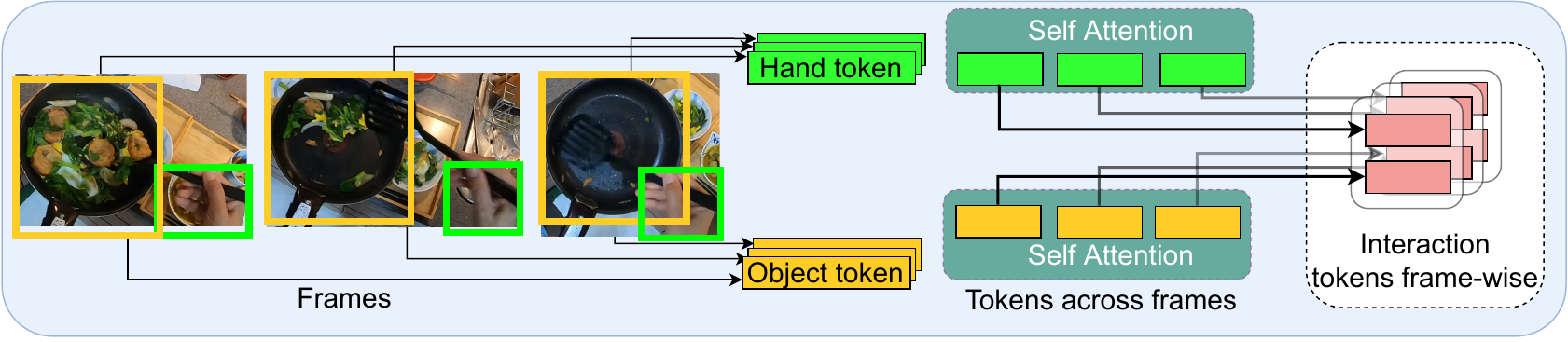}
    \caption{Modeling interaction tokens using Self-attention Over Time. We compute self-attention over hand tokens in all the frames to obtain refined hand tokens. We repeat this for every object region across frames. The refined hand and object tokens at every frame are the interaction tokens.}
    \label{fig:modelint_t}
\end{figure}

We model interaction tokens as the change in hands or objects individually over time as shown in \cref{fig:modelint_t} using self-attention.
Hand token $h_t$ is refined using only other hand tokens from all frames.
The query, key, and value are obtained from all hand tokens across all frames $\mathbf{q}_{h,t} = \mathbf{h}_{t} \mathbf{W}_{q},~\mathbf{k}_{h,t} = \mathbf{h}_{t} \mathbf{W}_{k},~\mathbf{v}_{h,t} = \mathbf{h}_{t} \mathbf{W}_{v}$ to obtain refined hand token $\tilde{\mathbf{h}}_t$ using self-attention.
%
We refine object tokens ${\mathbf{o}}_{i,t}$ of every object $i$ separately over frames using self-attention.
The query, key, and value for object $i$ are computed from its own tokens over all the frames $ \mathbf{q}_{i,t} = \mathbf{o}_{i,t} \mathbf{W}_{q},~~\mathbf{k}_{i,t} =  \mathbf{o}_{i,t} \mathbf{W}_{k},~~\mathbf{v}_{t} = \mathbf{o}_{i,t} \mathbf{W}_{v}$ to obtain refined object token
$\tilde{\mathbf{o}}_{i,t}$ using self-attention.
We call this method Self-attention Over time \textbf{SOT} and SOT interaction tokens $I_t = [\tilde{\mathbf{h}}_t,\tilde{\mathbf{o}}_{1,t}, \cdots, \tilde{\mathbf{o}}_{N,t}]$ consist of refined hand and object tokens.

\subsubsection{UB: Union Box of hand and nearest object}~\label{sec:ub}
We obtain the third type of interaction token using the hand and the nearest object in every frame.
We compute the union bounding box from the hand and the nearest object bounding boxes in every frame.
Here the hand and object union box is similar to the union of objects and human regions for interaction detection~\cite{li2021weakly}.
We then obtain the union tokens $\mathcal{U} \in \mathbb{R}^{T \times d}$ using the method described in \cref{sec:hotokens}.
Unlike the previous two approaches, the union region consists of both human and object features together.
We refine the $T$ union tokens in $\mathcal{U} \in \mathbb{R}^{T \times d}$ using self-attention to obtain interaction tokens $I_t\in \mathbb{R}^{T \times d}$.
We denote this approach as \emph{UB} for short.

%% file: model_3.tex
Next, we describe how to obtain context-infused interaction tokens in \cref{sec:tca}. 
Then, in \cref{sec:ant}, we describe how context-infused interactions are used to obtain interaction-centric video representation for action anticipation. 

\subsection{CI: Context-infused interaction tokens}\label{sec:tca}
\begin{figure}[t]
    \centering
    \includegraphics[width=\linewidth]{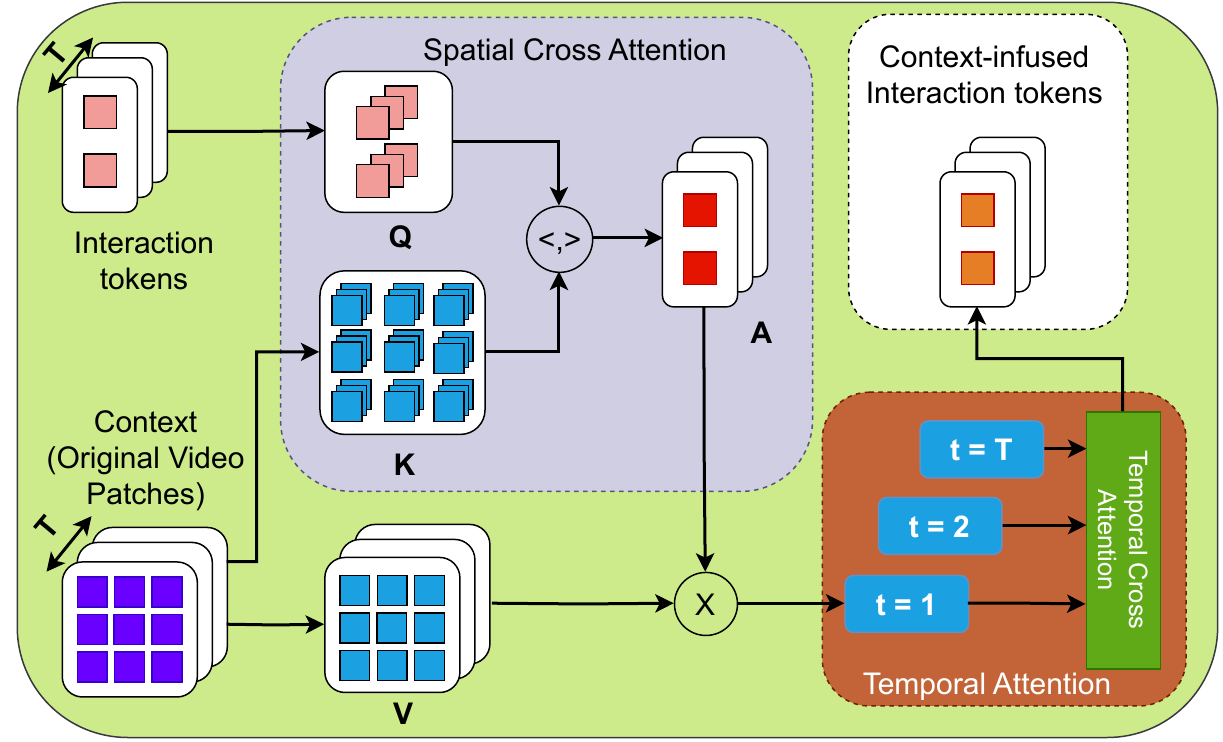}
    \caption{Context infusion into interaction region tokens using Trajectory Cross Attention. We compute spatial cross-attention (SCA) to find the best location for interaction trajectory by comparing the interaction query to context keys. Next, we pool the interaction trajectories across time to form connections across the interaction tokens in a frame.}
    \label{fig:tca}
\end{figure}

Here we discuss how we implement \cref{eq.tca.overview}.
The context plays an important role along with interaction in deciding what are the possible next actions.
For example, \textit{washing a plate} in \textit{kitchen sink} suggests that the next action is probably \textit{close the tap}.
Hence, we infuse context into interaction tokens by proposing Trajectory Cross Attention (TCA) based on trajectory attention~\cite{patrick2021keeping}. 
TCA maintains temporal correspondences between interaction tokens and context tokens of a  frame as shown in \cref{fig:tca}.

Our TCA formulation seeks the probabilistic path of an interaction token between frames. 
The interaction tokens act as the query on the video tokens $X$ that is representative of the context.
Let the video patch at spatial location $s$ in frame $t$ be given by $\mathbf{x}_{st} \in \mathbb{R}^d$. 
The key and value are obtained from the video patch. 
We have $N+1$ interaction tokens\footnote{1 in case of UB interaction modeling} in every frame.
For each interaction token $\mathbf{y}_t \in I_t$, we obtain a set of trajectory tokens $\hat{\mathbf{y}}_{tt'} \in \mathbb{R}^d, \forall t' \geq t$ that represents pooled information weighted by the trajectory probability.
The pooling operation implicitly looks for the best location $s$ at frame $t' \geq t$ by comparing the interaction query $\mathbf{q}_{t} = \mathbf{y}_t \mathbf{W}_q$ to the context keys $\mathbf{k}_{st'} = \mathbf{x}_{st'} \mathbf{W}_k$ using $q,k,v$-attention. 
%
Attention is applied spatially and independently for all the interaction tokens in every frame.
This is complementary to our previously computed cross attention (\cref{hchange} and \cref{ochange}) where hand/object query tokens are refined with respect to other object/hand key and value token. 
In TCA, we seek to infuse interaction tokens with the visual context of the video which is similar to \cite{jaegle2021perceiver} where cross-attention is used to infuse query tokens with information from key and value tokens.

Once trajectories are computed, we pool them across time to reason about connections across the interaction regions in a frame given the environment. 
For temporal pooling, the trajectory tokens are projected to a new set of queries, keys, and values $\hat{\mathbf{q}}_{tt'} =  \hat{\mathbf{y}}_{tt'} \mathbf{W}_q,~ \hat{\mathbf{k}}_{tt'}  = \hat{\mathbf{y}}_{tt'} \mathbf{W}_k,~ \hat{\mathbf{v}}_{tt'} = \hat{\mathbf{y}}_{tt'} \mathbf{W}_v$, respectively.
The new query $\hat{\mathbf{q}}_{tt'}$ has information across the entire trajectory that extends across the entire observed video frames.  
We perform temporal pooling using 1D attention across the new time (trajectory) dimension to obtain refined interaction tokens.
We term these refined interaction tokens as \emph{context-infused interaction tokens} $\tilde{I} = [\mathbf{\tilde{y}}_{1}, \cdots,  \mathbf{\tilde{y}}_{T} ]$.

\subsection{ICV: Interaction-centric Video Representation}\label{sec:ant}
Here we discuss how we implement \cref{vchange}.
The next action is dependent on interactions and context. 
Therefore, our goal is to obtain a video representation that incorporates information from the context-infused interaction tokens. 
We concatenate the video tokens $X$ and the context-infused interaction tokens $\tilde{I}$ to form the augmented video representation $X_{a}$\footnote{$X_a \in \mathbb{R}^{T\times (H\times W + N+1)\times d}$ (for SCA and SOT) or $X_a \in \mathbb{R}^{T\times (H\times W + 1)\times d}$ for UB}.
Then we perform self-attention over the augmented video tokens $\mathbf{x}^a_{r} \in X_a$ wherein video tokens attend over the interaction regions tokens and vice-versa.
After self-attention, we obtain refined augmented video tokens $\tilde{\mathbf{x}}^a_{r}$. 
We construct the interaction-centric video representation $X_I$ using $\tilde{\mathbf{x}}^a_{r}$ that corresponds to the original video tokens. 
\begin{align}
X_I = \tilde{\mathbf{x}}^a_{r}, \forall {r}, \text{ if } \mathbf{x}^a_{r} \in X.
\end{align}
We call $X_I$ as interaction-centric video representation in the same vein as object-centric video representation~\cite{herzig2022object}.


Till now, we have only applied a single attention layer for interaction modeling, context infusion for interaction, and interaction-centric video representation. 
In literature, video transformer approaches~\cite{patrick2021keeping,ou2022object,arnab2021vivit} have shown excellent performance with multiple layers of attention.
We apply 12 layers of Trajectory Attention following MotionFormer~\cite{patrick2021keeping} on the interaction-centric video representation to obtain the final video representation $\tilde{X}_I$.
The reason for choosing MotionFormer is that it performs best empirically.
The classification token $\mathbf{x}_{cls}$ obtained at the end of multiple trajectory attention layers is used to predict the next action, i.e., $\hat{\mathbf{a}}_{next} = \phi(\tilde{\mathbf{x}}_{cls})$ where $\phi$ is a linear layer.

\textbf{Loss function.} We use cross-entropy loss for the next action label to train our model. 
We compare the model's prediction $\hat{\mathbf{a}}_{next} $ with the ground truth one-hot label $\mathbf{a}_{next}$ for the next action as follows
\begin{align}
\mathcal{L}_{ant} = - \sum  \mathbf{a}_{next} \odot \log(\hat{\mathbf{a}}_{next}).
\label{cel}
\end{align}
It should be noted that as we train our model with the cross-entropy loss to predict the next action, the interaction tokens are optimized for finding the most influential interactions when predicting the next action (see \cref{fig:qual}).

%% file: exp_new.tex

\subsection{Datasets and Implementation Details}
We evaluate and compare our methods on two large unscripted action anticipation datasets \textit{EPIC-KITCHENS100}\cite{damen2022rescaling} (EK100) and \textit{EGTEA Gaze+}\cite{min2021integrating}.
For EK100, we report results on the test set evaluation server that uses mean-top5 recall as the metric.
For EK100 and EGTEA, anticipation gap ($T_a$) is 1s and 0.5s, respectively.

We follow MotionFormer~\cite{patrick2021keeping} and use 16-frame long clips of resolution $224\times224$ uniformly sampled from an observed video of 64 frames ( approximately 2s).
Every 3D video token is extracted from a video patch of size $2 \times 16 \times 16$. 
We extract hand, object, and union tokens following the strategy explained in \cref{sec:hotokens}.
Then, to implement InAViT(SCA) (\cref{sec:sca}), we use a single cross-attention layer with 12 attention heads.
Similarly, InAViT(SOT) and InAViT(UB) are implemented with a self-attention layer with 12 heads.
We set the number of objects per frame as 4 for EK100 and 2 for EGTEA based on empirical performance (this also makes sure batch processing is efficient). 
We report the results of varying the number of objects per frame in Appendix C. 
If there are fewer objects (less than 4 or 2 respectively), then we zero pad them using null tokens and mask them, which will not impact the model.
The same configuration of objects is used for training other baseline models such as ORViT-MF~\cite{herzig2022object}.

EK100 provided hand detections do not contain hand annotations for 20\% of the frames.
Since our aim is not localize hands accurately, we reduce the threshold to 0.05 from 0.1 used by EK100 to get hand detections in all frames.
Lowering the threshold introduces bounding-boxes that cover the region around the hand that is useful for interaction region modeling and anticipation as shown in \cref{fig:det_predict}.
The number of hand regions per-frame is set to 1 as both hands are not visible in most frames. 
If there are two hands in a frame, we randomly pick one and track it using SORT. Some analysis and statistics about detected objects in frames are shown in Appendix B.

We use one layer of trajectory cross-attention with 12 attention heads and a temporal resolution of 8 to obtain the context infusion of interaction tokens (\cref{sec:tca}).
We then concatenate the refined interaction and original video tokens to form the augmented video tokens. 
We use a single self-attention layer with 12 heads on the augmented video tokens to obtain interaction-centric video tokens (\cref{sec:ant}).
Finally, we apply MotionFormer on the interaction-centric video tokens to predict the next action (\cref{sec:ant}). 
We use a batch size of 16 video(clips) to train on 4 RTX A5000 GPUs with 24 GB memory each and the learning rate is set to $1e^{-4}$ with AdamW optimizer. We will release our code.

\subsection{Ablation on InAViT}
\textbf{Component-wise validation.}
In \cref{table:ablationtable}(a), we show the contribution of each component of InAViT - interaction modeling using SCA (\cref{ochange,hchange}), \textbf{C}ontext \textbf{I}nfusion of interaction tokens (\textbf{CI}) (\cref{eq.tca.overview}), and \textbf{I}nteraction-\textbf{C}entric \textbf{V}ideo representation using trajectory attention (\textbf{ICV}) (\cref{vchange}).
Context infusion improves overall performance but we see the biggest improvement when we add the interaction-centric video representation (i.e. \cref{vchange}). 
So, we conclude that interactions and the interaction centric video representations are important for action anticipation.

\textbf{Comparing different interaction regions.} 
In \cref{table:ablationtable}(b), we compare the three models for interaction region modeling - spatial cross attention (SCA+ CI + ICV), spatial attention over time (SOT+ CI + ICV), and union boxes (UB + CI + ICV)  described in \cref{sec:modelint}. 
In our comparison, SCA performs the best on overall, unseen, and tail classes compared to both SOT and UB.
SCA contextualizes the visual change of each hand/object better using other objects compared to SOT which computes visual change individually.
UB's focus is narrower than SCA as it only considers the nearest object and potentially leaves out other objects that can be used in the next action.
SCA performs much better in tail classes where few examples are available and the model relies on visual information for anticipation.
As SCA uses all the objects to model interactions, it extracts the most visual information from every frame to make better predictions.

\textbf{Only Hand/Object as interaction regions.}
The best-performing SCA interaction model involves both refined hand and refined object tokens as interaction region tokens.
In \cref{table:ablationtable}(c), we compare the contribution of only refined hand (SCA(Hand)+CI + ICV) and only object tokens (SCA(Obj)+CI + ICV) by using either of them as interaction region tokens.
As in SCA formulation, we refine hand tokens with objects and object tokens with hand and other objects.
Refined hand tokens perform better than refined object tokens as the position of the hand is vital in determining what object(s) can be used next.
Still, interaction region tokens containing both refined hands and object tokens (SCA+ CI + ICV) perform the best.
We conclude that modeling the changes in hand and object tokens provides useful information to improve action anticipation.
\begin{table}[t!]
\centering
\begin{tabular}{lccc} 
\hline
\multirow{2}{*}{Method} & Overall & Unseen  & Tail \\ 
 & Action(\%) & Action(\%) & Action(\%) \\ 
\hline
SCA & 12.66 & 15.49 & 06.03 \\ 
SCA + CI & 14.21 & 14.26 & 09.12 \\ 
SCA+ICV & 22.21 & 20.85 & 17.07 \\ 

\begin{tabular}[c]{@{}l@{}}SCA + CI + ICV\end{tabular} & \textbf{23.75} & \textbf{23.49} & \textbf{18.11} \\ \hline

\multicolumn{4}{c}{(a) Component-wise validation of InAViT} \\
\hline
UB+CI + ICV & 22.75 & 22.14 & 17.04 \\ 
SOT+CI + ICV & 22.48 & 20.56 & 17.46 \\ 
\hline
\multicolumn{4}{c}{(b) Comparing interaction modeling methods} \\ 
\hline
SCA-(Hand) + CI + ICV & 23.27 & 23.21 & 17.57 \\
SCA-(Obj) + CI + ICV & 22.49 & 22.23 & 16.73 \\ 
\hline
\multicolumn{4}{c}{(c) Comparing refined hand vs. object as interaction tokens} \\ 
\end{tabular}
\caption{Ablation of InaViT on EK100 evaluation server [\textbf{Test set}]. Verb and Noun results are in Appendix.}
\label{table:ablationtable}
\end{table}
\begin{table}[t!]
\centering
\begin{tabular}{lccc}
\hline
\multirow{2}{*}{Method} & Overall & Unseen  & Tail \\ 
 & Action(\%) & Action(\%) & Action(\%) \\ 
\hline
\begin{tabular}[c]{@{}l@{}}SCA + CI + ICV\end{tabular} & \textbf{23.75} & \textbf{23.49} & \textbf{18.11} \\
SCA + CI(Mask FG) + ICV & 08.05 & 05.92 & 05.84 \\ 
SCA+ Concat + ICV & 22.14 & 23.47 & 17.24 \\
\hline
\end{tabular}
\caption{Effect of infusing context in different ways. CI(Mask FG):  Context Infusion with foreground hands and objects masked out, Concat: Context infusion by concatenating context tokens with interaction tokens. Results are on evaluation server [\textbf{Test set}].}
\label{table:context}
\end{table}

\textbf{Effect of infusing context.}
We evaluate the effect of context infusion on interaction tokens in \cref{table:context}. 
For this comparison, we change either the input or mechanism of context infusion(\cref{sec:tca}).
Interaction modeling is done using SCA and interaction-centric video representation is computed using the original video tokens.
For CI (Mask FG), we mask the foreground i.e., hands and objects from the context.
We crop the objects and hands based on the bounding boxes and apply a Gaussian filter to soften the edges.
We call this masked foreground context and use it to refine interaction tokens.
The performance of (SCA+CI(Mask FG)+ICV) is quite poor which means that the complete context with foreground objects (SCA+CI+ICV) is better for refinement.
We also find that the concatenation of context (video tokens) to interaction tokens (SCA+Concat+ICV) is worse than our proposed  (SCA+CI+ICV) approach.
\subsection{Comparison with state-of-the-art}
We now compare our best-performing InAViT (SCA+CI+ICV) model against state-of-the-art approaches.
On EK100 evaluation server, InAViT significantly outperforms other approaches as seen in \cref{tab:sota_ek100}.
InAViT's performance is much better than AVT \cite{girdhar2021anticipative},  MeMViT \cite{wu2022memvit}, and RAFTformer \cite{girase2023latency} on EK100 which also use visual transformers for representing the video.
We also compare InAViT against the baseline MotionFormer (MF)~\cite{patrick2021keeping} and object-centric video representation ORViT-MF~\cite{herzig2022object}.
As ORViT and MF are not trained for action anticipation, we train them using the official repositories.

We are the first to show the effectiveness of MF and ORVIT-MF for action anticipation which alone outperforms prior state-of-the-art methods.
Our approach InAViT performs even better than both MF and ORVIT-MF and achieves significantly better results than the previous best results, the Abstract Goal \cite{roy2022predicting} on EGTEA (\cref{tab:sota_egtea}) and AVT \cite{girdhar2021anticipative} on EK100.
In fact, InAViT outperforms \cite{roy2022predicting} by 18\% in mean accuracy and 20.8\% in top-1 accuracy on EGTEA.
It also outperforms the human-object interaction method in \cite{liu2020forecasting} by 30\% and 35\% on mean and top-1 accuracy, respectively.
Similarly, InAViT outperforms  AVT \cite{girdhar2021anticipative} by 22\%, 17\%, and 7\%, in the overall verb, noun, and action anticipation on EK100 which is impressive given the large number of actions (3805), nouns (300), and verbs (97).
On EK100, InAViT outperforms AVT on unseen and tail action anticipation by 12\% and 4\%, respectively.  This demonstrates that InAViT is effective at predicting rare actions and generalizes much better to new environments. 

\begin{table}[t]
\centering
\resizebox{\linewidth}{!}{
\begin{tabular}{llccccccccc} 
\hline
\multirow{2}{*}{Set}& \multirow{2}{*}{Method} & \multicolumn{3}{c}{Overall (\%)} & \multicolumn{3}{c}{Unseen (\%)} & \multicolumn{3}{c}{Tail (\%)} \\ 
\cline{3-11}
 & &  Verb & Noun & Action & Verb & Noun & Action & Verb & Noun & Action \\ 
\hline
\multirow{7}{*}{\begin{tabular}[c]{@{}l@{}}Val\end{tabular}}& {RU-LSTM~\cite{damen2022rescaling}}  &  23.20 & 31.40 & 14.70 & 28.00 & 26.20 & 14.50 & 14.50 & 22.50 & 11.80 \\ 
& {T. Agg.~\cite{sener2020temporal}} & 27.80 & 30.80 & 14.00 & 28.80 & 27.20 & 14.20 & 19.80 & 22.00 & 11.10\\ 
& AFFT \cite{zhong2023anticipative} & 22.80 & 34.60 & 18.50 & 24.80 & 26.40 & 15.50 & 15.00 & 27.70 & 16.20 \\ 
& AVT~\cite{girdhar2021anticipative} & 28.20 & 32.00 & 15.90 & 29.50 & 23.90 & 11.90 & 21.10 & 25.80 & 14.10 \\ 
& TrAc~\cite{gu2021transaction} & 35.04 & 35.49 & 16.60 & 34.64 & 27.26 & 13.83 & 30.08 & 33.64 & 15.53  \\ 
& MeMViT~\cite{wu2022memvit} & 32.20 & 37.00 & 17.70 & 28.60 & 27.40  & 15.20 & 25.30 &  31.00 & 15.50 \\ 
& Rformer \cite{girase2023latency} & 33.80 & 37.90 & 19.10 & - & - & - & - & - & - \\  
& MF$^{*}$ & 47.14 & 46.92 & 21.52 & 42.33 & 47.22 & 21.99 & 40.47 & 38.66 & 16.36 \\ 
& ORViT-MF$^{*}$ & 46.71 & 47.91 & 23.54 & 38.99 & 45.32 & 23.60 & 37.28 & 37.81 & 17.10 \\ \cline{2-11}
& InAViT (Ours) & \textbf{52.54} & \textbf{51.93} & \textbf{25.89} & \textbf{46.45} & \textbf{51.30} & \textbf{25.33} & \textbf{45.34} & \textbf{39.21} & \textbf{20.22} \\ \hline
\\ \hline

\multirow{7}{*}{\begin{tabular}[c]{@{}l@{}} Test \end{tabular}}& {RU-LSTM~\cite{damen2022rescaling}}  & 25.25 & 26.69 &	11.19 & 19.36 & 26.87 & 09.65 & 17.56 &	15.97 &	07.92 \\ 
& {T. Agg.~\cite{sener2020temporal}} &
21.76 &	30.59 &	12.55 &	17.86 &	27.04 &	10.46 &	13.59 &	20.62 &	08.85 \\ 
& AFFT \cite{zhong2023anticipative} & 20.70 & 31.80 & 14.90 & 16.20 & 27.70 & 12.10 & 13.40 & 23.80 & 11.80 \\ 
& AVT~\cite{girdhar2021anticipative} & 26.69 & 32.33 & 16.74 & 21.03 & 27.64 & 12.89 & 19.28 &	24.03 &	13.81 \\ 
& Abs. goal~\cite{roy2022predicting} & 31.40 & 30.10 & 14.29 & 31.36 & 35.56 & 17.34 & 22.90 & 16.42 & 07.70 \\ 
& TrAct~\cite{gu2021transaction} & 36.15 & 32.20 & 13.39 & 27.60 & 24.24 & 10.05 & 32.06 & 29.87 & 11.88 \\ 
& Rformer \cite{girase2023latency} & 30.10 & 34.10 & 15.40 & - & - & - & - & - & - \\ 
& MF$^{*}$ & 45.14 & 45.97 & 19.75 & 40.36 & 45.28 & 19.49 & 39.17 & 35.91 &	14.11 \\ 
& ORViT-MF$^{*}$ & 43.74 &	46.61 & 21.53 & 38.99 & 45.32 & 21.47 & 37.28 & 35.78 & 15.96 \\ \cline{2-11}
& InAViT (Ours) & \textbf{49.14} & \textbf{49.97} & \textbf{23.75} & \textbf{44.36} & \textbf{49.28} & \textbf{23.49} & \textbf{43.17} & \textbf{39.91} & \textbf{18.11} \\ \hline
\end{tabular}
}
\caption{Comparison with state-of-the-art on EK100 validation (Val) and evaluation server (Test). 
InAViT significantly outperforms other Transformer-based approaches such as AVT, MotionFormer and object-centric Motionformer (ORViT-MF). 
$^*$We trained MF and ORViT-MF for action anticipation using their official repositories.}
\label{tab:sota_ek100}
\end{table}

\begin{table}
\centering
\begin{tabular}{lcccccc} 
\hline
\multirow{2}{*}{Method} & \multicolumn{3}{c}{Top-1 Accuracy (\%)} & \multicolumn{3}{c}{Mean Class Accuracy (\%)} \\ 
\cline{2-7}
 & VERB & NOUN & ACT. & VERB & NOUN & ACT. \\ 
\hline
FHOI~\cite{liu2020forecasting} & 49.0 & 45.5 & 36.6 & 32.5 & 32.7 & 25.3 \\ 

AFFT \cite{zhong2023anticipative} & 53.4 & 50.4 & 42.5 & 42.4 & 44.5 & 35.2  \\ 

AVT~\cite{girdhar2021anticipative} & 54.9 & 52.2 & 43.0 & 49.9 & 48.3 & 35.2 \\ 
Abs. Goal~\cite{roy2022predicting} & 64.8 & 65.3 & 49.8 & 63.4 & 55.6 & 37.4\\

MF$^{*}$ & 77.8 & 75.6 & 66.6 & 77.5 & 72.1 & 56.9 \\ 
ORVIT-MF$^{*}$ & 78.8 & 76.3 & 67.3 & 78.8 & 75.8 & 57.2  \\ 
\hline
InAViT (Ours) & \textbf{79.3} & \textbf{77.6} & \textbf{67.8} & \textbf{79.2} & \textbf{76.9} & \textbf{58.2} \\ \hline
\end{tabular}
\caption{Comparison of anticipation performance on EGTEA Gaze+. Complete table in Appendix.}
\label{tab:sota_egtea}
\end{table}

In \cref{egteavarant}, we vary the anticipation gap ($T_a$) for EGTEA dataset. 
InAViT is more robust for longer anticipation because it can make use of additional human-object interaction information than ORVIT-MF.
We also analyzed the computational cost and number of parameters of InAViT and found that the baseline MotionFormer requires 370 GFlops with 143.9M parameters, while InAViT requires 391 GFlops with 157.2M parameters and ORVIT-MF requires 403 GFlops with 172.1M parameters.
Hence, InAViT is less computationally expensive and has lower number of parameters than ORVIT-MF and it still performs better on action anticipation. 
\begin{table}
\centering
\scriptsize
\begin{tabular}{l|cccc|c|c} 
\hline
\multirow{2}{*}{\begin{tabular}[c]{@{}l@{}}Top-1 \\Action Acc.\end{tabular}} & \multicolumn{4}{c|}{Anticipation gap ($T_a$ in s)} & GFlops & Params\\ 
\cline{2-5} 
& 2 & 1.5 & 1 & 0.5 & & \\ 
\hline
MF & 60.2 & 61.7 & 64.2 & 66.6 & 370 & 143.9M\\
ORVIT-MF & 62.2 & 63.6 & 66.1 & 67.3 & 403 & 172.1M \\
InAViT & \textbf{64.1} & \textbf{65.8} & \textbf{66.9} & \textbf{67.8} & 391 & 157.2M \\
\hline
\end{tabular}
\caption{InAViT performs better even on larger anticipation gap of 1.5 and 2s (EGTEA). InAViT has better performance than ORVIT with lower computations and parameters.}
\label{egteavarant}
\end{table}

\begin{figure}
\begin{tabular}{c}
\includegraphics[width=0.9\linewidth,clip=True]{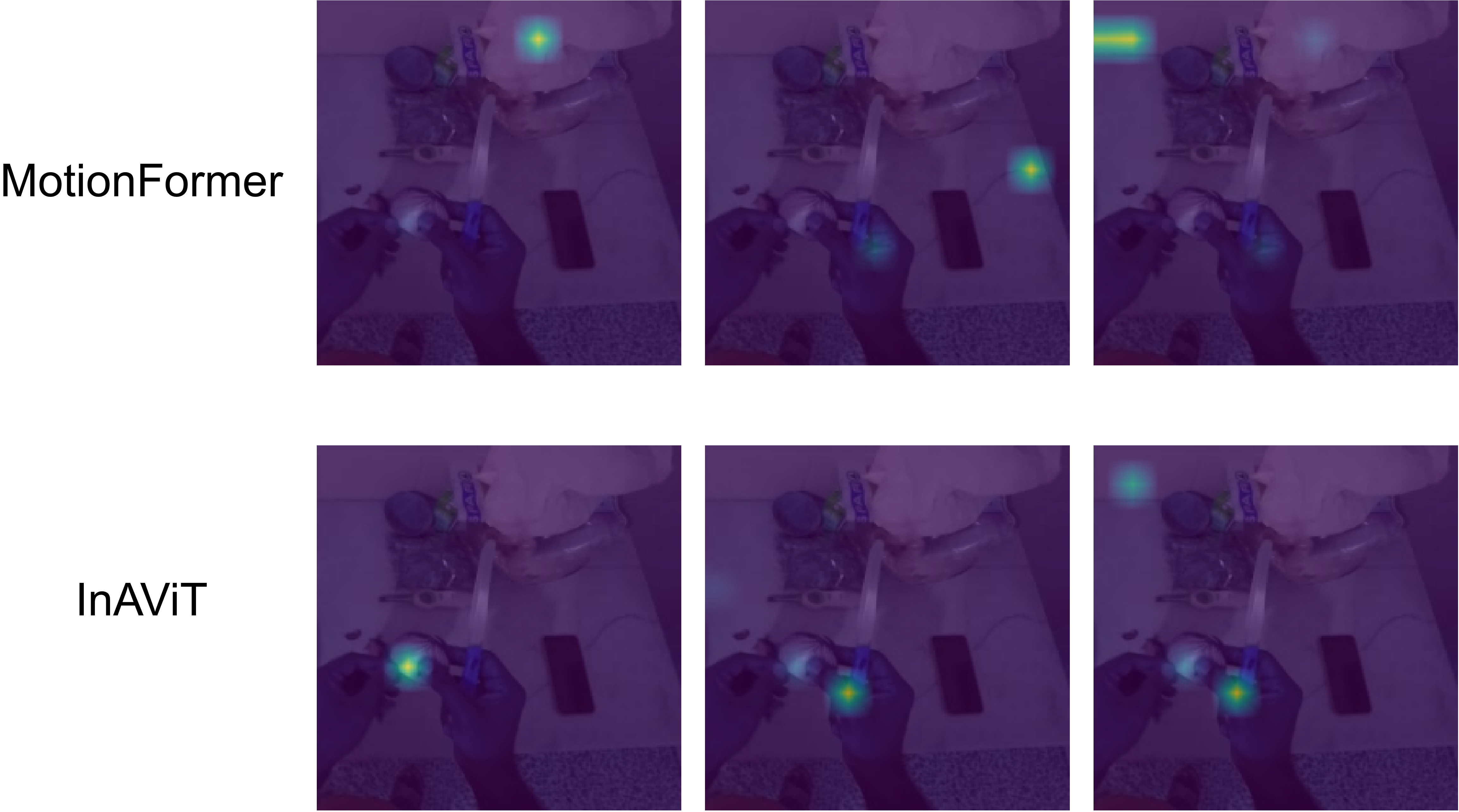}\\
(a) Next action: \textit{peel onion} \\
\includegraphics[width=0.9\linewidth,clip=True]{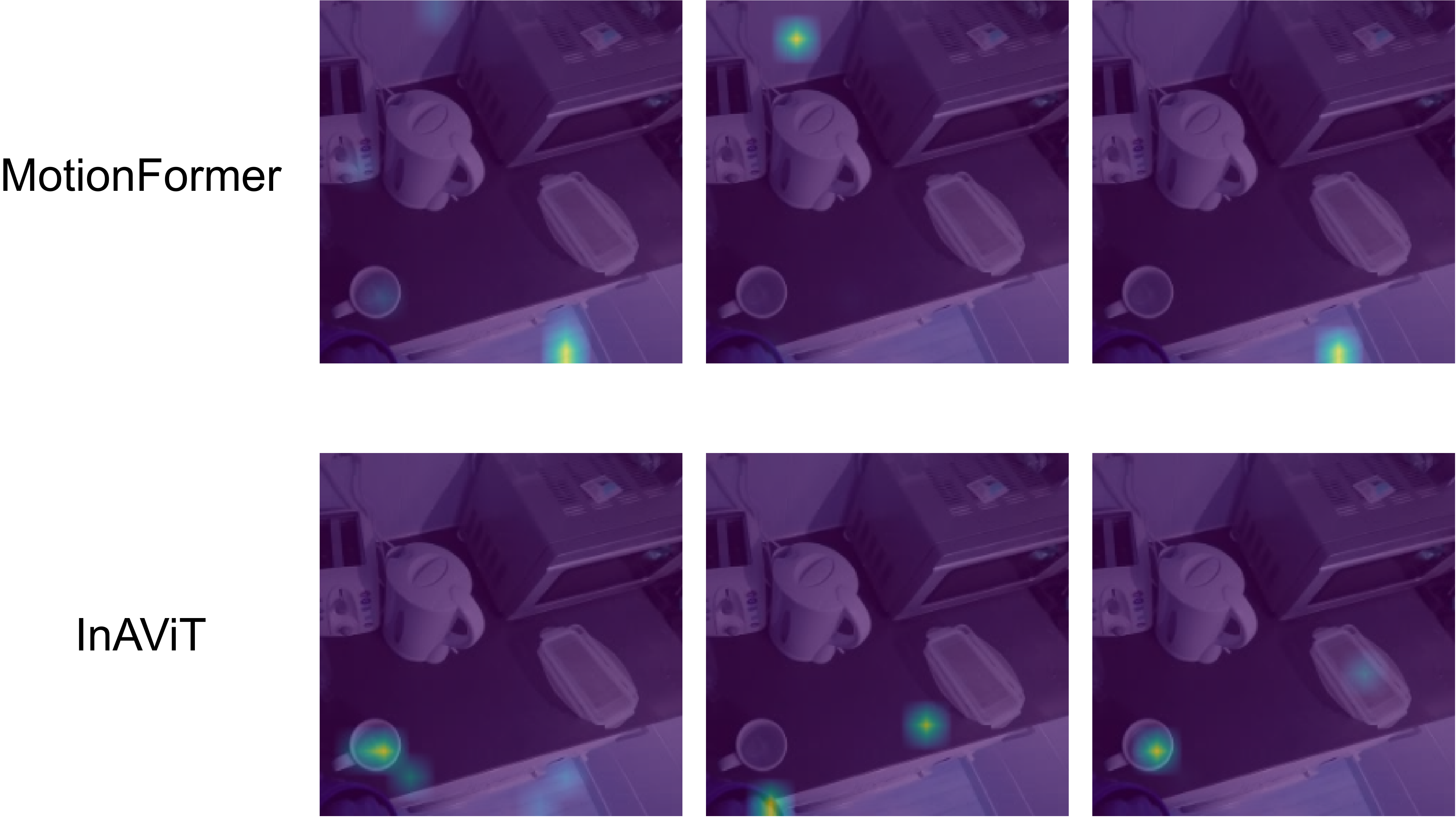}\\
(b) Next action: \textit{pour sugar}
\end{tabular}
    \caption{InAViT attends to the location(s) where the next action will occur in (a) onion and (b) cup and sugar.}
    \label{fig:qual}
\end{figure}
\begin{figure}[!ht]
     \centering
     \begin{subfigure}[b]{\linewidth}
         \centering
         \includegraphics[width=\linewidth]{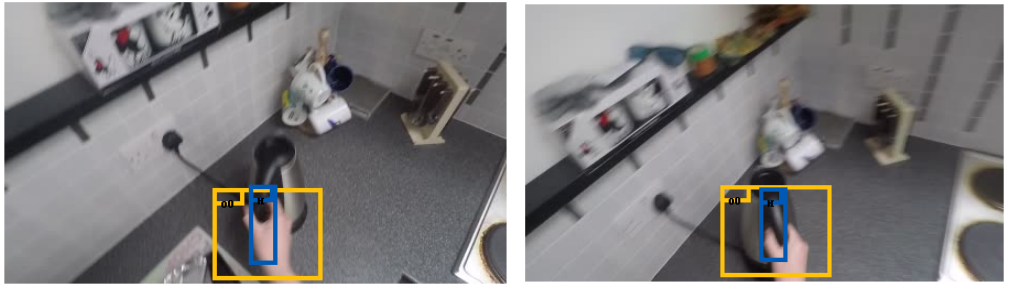}
         \caption{GT: \textit{fill kettle}, Pred: \textit{\textcolor{green}{fill kettle}}}
         \label{fig:five over x}
     \end{subfigure}
     \begin{subfigure}[b]{\linewidth}
         \centering
         \includegraphics[width=\linewidth]{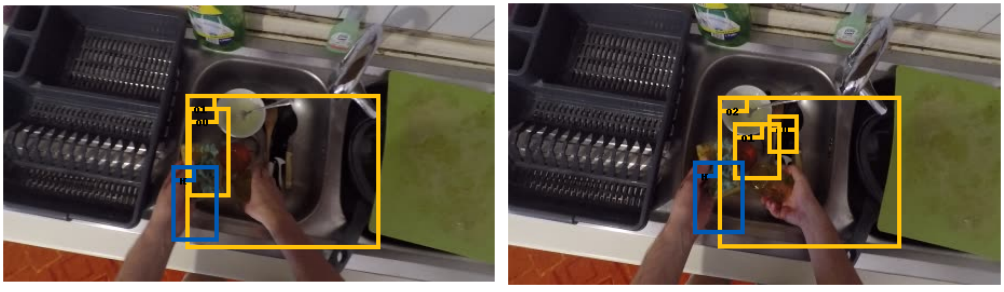}
         \caption{GT: \textit{wash glass}, Pred: \textit{\textcolor{green}{wash glass}}}
         \label{fig:five over x}
     \end{subfigure}
     
     \begin{subfigure}[b]{\linewidth}
         \centering
         \includegraphics[width=\linewidth]{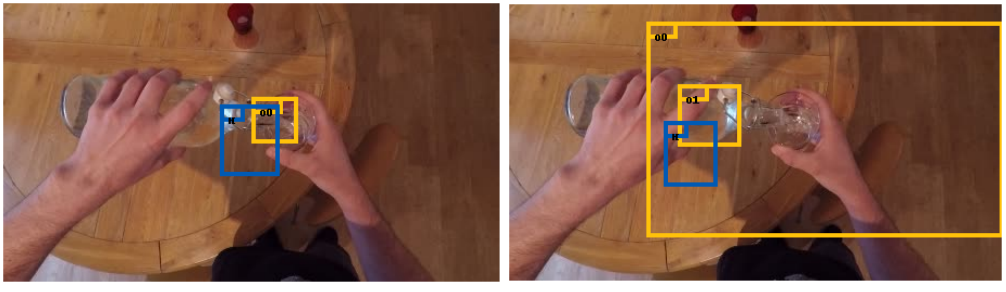}
         \caption{GT: \textit{drink water}, Pred: \textit{\textcolor{green}{drink water}}}
         \label{fig:five over x}
     \end{subfigure}
     \begin{subfigure}[b]{\linewidth}
         \centering
         \includegraphics[width=\linewidth]{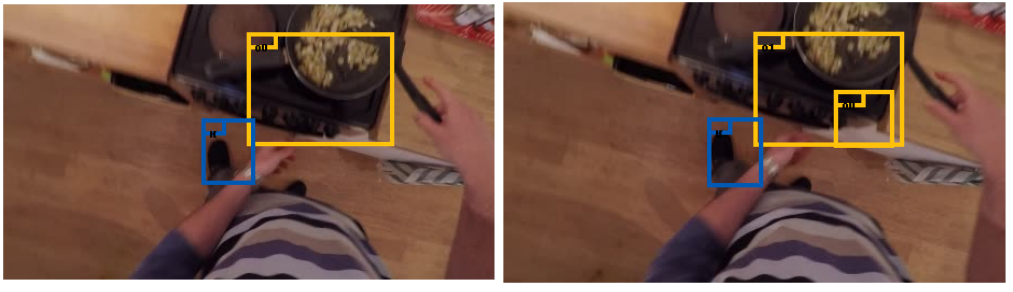}
         \caption{GT: \textit{adjust hob}, Pred: \textit{\textcolor{green}{adjust hob}}}
         \label{fig:five over x}
     \end{subfigure}
        \caption{Some examples of InAViT's anticipation where \textcolor{green}{green} is correct and \textcolor{red}{red} is incorrect.  InaViT can anticipate correctly with hand detections (in \textcolor{blue}{blue}) that precisely capture the hand (a)(b) or cover the region around the hand in (c)(d). }
        \label{fig:det_predict}
\end{figure}

\textbf{Qualitative Results.} In \cref{fig:qual}, we visualize the attention map of the $\mathbf{x}_{cls}$ token used for action anticipation on all spatial tokens across the frames. 
This helps us understand where InAViT focuses compared to MotionFormer. 
Motionformer attention is divided into many areas while InAViT attends to the interaction.
While anticipating \textit{peel onion} (\cref{fig:qual}(a)), InAViT pays high attention to the exact location the onion is being peeled.
Similarly, when anticipating \textit{pour sugar} (\cref{fig:qual}(b)), InAViT attends to both the cup and the sugar container. 
The frames for visualization are chosen based on the significant motion of hands and objects during the observed action.
In \cref{fig:det_predict}(c)(d), we show that InAViT anticipates the action correctly even if the bounding box covers the region around the hand.
We show more qualitative results in Appendix.

%% file: conclusion.tex

We present a novel hand-object interaction modeling visual transformer InAViT for egocentric action anticipation.
We show that refinement of hand and object boxes with respect to each other i.e. SCA is a stronger prior for interaction modeling than isolated refinement as in SOT (Self-attention of hand/object Over Time).
We also find that union box comprising of both hands and objects is ineffective at interaction modeling compared to mutual refinement of individual hand and object boxes.
We show that incorporating environment information makes interactions more effective at action anticipation.
Particularly, using Trajectory Cross Attention for environment-refinement of interaction tokens is better than concatenation.
Finally, we show that interaction-centric video representation is better at anticipation than object-centric video representation.

%% file: sup.tex
\appendix
\section{Cross-Attention and Self-Attention}~\label{sec:defatt}
In cross-attention, queries are obtained from the target that needs to be refined.
The key and values are obtained from the source that needs to be queried to obtain the refined target tokens. 
Let $\mathbf{a}_m \in A$ be a target token from all target tokens $A$ and $\mathbf{b}_n \in B$ be a source token from all the source tokens $B$.
The queries, keys, and values are constructed as
\begin{align}
 \mathbf{q} = \mathbf{a}_m \mathbf{W}_q, \mathbf{k} = \mathbf{b}_n \mathbf{W}_k, \mathbf{v} = \mathbf{b}_n \mathbf{W}_v 
\end{align}
where $\mathbf{W}_q, \mathbf{W}_k, \mathbf{W}_v$ are learnable weights.
In the rest of the paper, we use $\mathbf{W}_q,~\mathbf{W}_k,\text{ and }~\mathbf{W}_v$, as learnable weights for query, key, and value, respectively, for all attention operations.
We do this for clarity in notation but all the weights are different and learned independently. 
In self-attention, query, key, and value are all obtained from target tokens $\mathbf{a}_m \in A$.
\begin{align}
 \mathbf{q} = \mathbf{a}_m \mathbf{W}_q, \mathbf{k} = \mathbf{a}_m \mathbf{W}_k, \mathbf{v} = \mathbf{a}_m \mathbf{W}_v 
\end{align}
The output of both cross-attention and self-attention is the refined target token  $\tilde{\mathbf{a}}_m$.
\begin{align} \label{eq:att}
 \tilde{\mathbf{a}}_m = \sum \mathbf{v} \frac{\text{exp}\left< \mathbf{q}, \mathbf{k}\right>}{\sum {\text{exp}\left< \mathbf{q}, \mathbf{k} \right>}} 
\end{align}
We omit the factor $d^{1/2}$ for clarity and assume that both queries and keys have been scaled by $d^{1/4}$~\cite{vaswani2017attention}.

\section{Does failed hand detection affect performance?}

\begin{figure}
    \centering
    \begin{subfigure}{0.6\linewidth}
    \includegraphics[width=\linewidth]{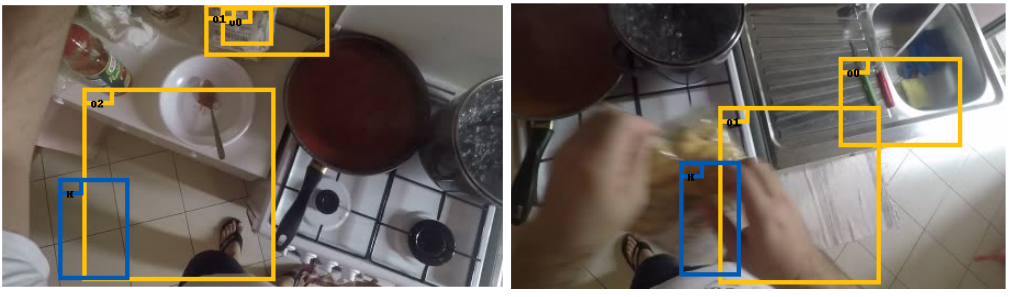}
    \caption{GT: \textit{open pasta}. Pred: \textit{\textcolor{green}{open pasta}}}
    \end{subfigure}
    \begin{subfigure}{0.6\linewidth}
    \includegraphics[width=\linewidth]{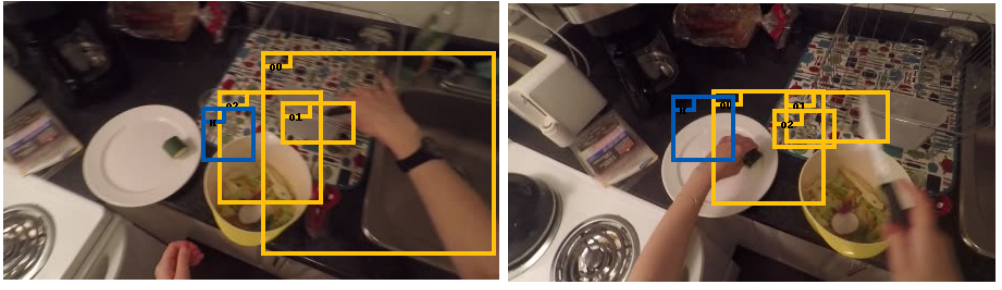}
    \caption{GT: \textit{cut cucumber}. Pred: \textit{\textcolor{green}{cut cucumber}}}
    \end{subfigure}
    \begin{subfigure}{0.6\linewidth}
    \includegraphics[width=\linewidth]{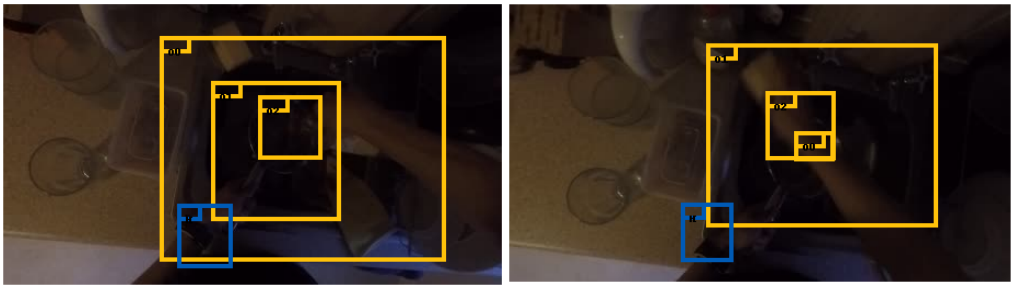}
    \caption{GT: \textit{wash pan}. Pred: \textit{\textcolor{green}{wash pan}}}
    \end{subfigure}
     \begin{subfigure}[b]{0.6\linewidth}
     \centering
     \includegraphics[width=\linewidth]{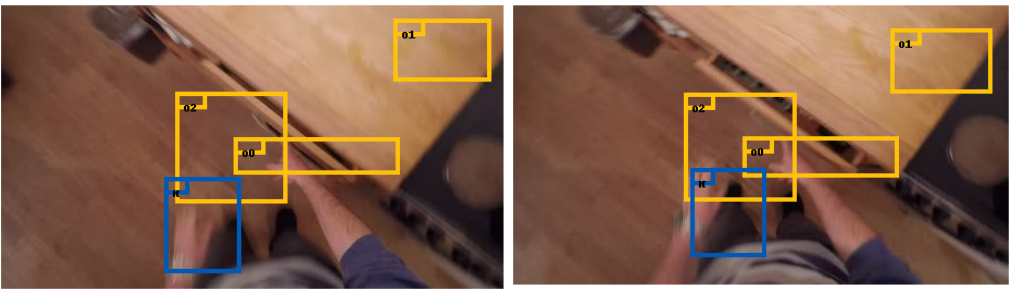}
     \caption{GT: \textit{take knife}, Pred: \textit{\textcolor{green}{take knife}}}
     \label{fig:five over x}
    \end{subfigure}
    \begin{subfigure}[b]{0.6\linewidth}
     \centering
     \includegraphics[width=\linewidth]{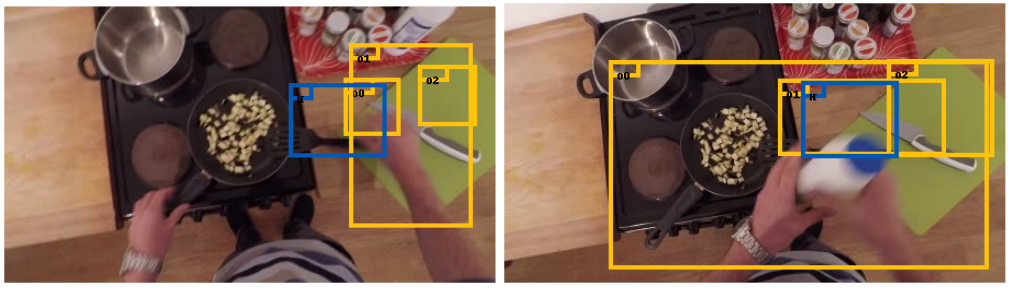}
     \caption{GT: \textit{pour salt}, Predicted: \textit{\textcolor{green}{pour salt}}}
     \label{fig:y equals x}
    \end{subfigure}
    \caption{More examples of detections that cover areas around the hand after lowering threshold of Faster RCNN to 0.05 on EK100. The hand detections (in blue) along with the objects (in yellow) cover the region around the hand that is useful for interaction. InAViT still anticipates correctly with noisy hand detection.}
    \label{fig:detections.all}
\end{figure}

No, this is not an issue.
The below Table~\ref{tab:hands} shows the number of frames with no hands is less than 0.5\% of all frames. Hence, the influence of missing hands is negligible. 
We use the same hand-object detector used by the EK100 dataset. We use a threshold of 0.05 instead of 0.1  to obtain more hand detections.
Object occlusion (50\% or more) due to hands happens only in 8.6\% of the frames in EK100.
In \cref{fig:detections.all}, we show that reducing the threshold for hand detection results results in noisy hand detections but InAViT is still able to predict method is able to predict the future action correctly.

\begin{table}[h!]
\centering
\scriptsize
\resizebox{0.8\linewidth}{!}{
\begin{tabular}{l|c|c|c|c|c|c}
\hline
\multirow{3}{*}{\textbf{Split}} & \multicolumn{6}{c}{\# of hands present per frame(\%)} \\ \cline{2-7}
&
  \multicolumn{3}{c|}{EK100} &
  \multicolumn{3}{c}{EGTEA} \\ \cline{2-7} 
      & 0 & 1 & 2 & 0 & 1 & 2 \\ \hline
train & 0.18 & 05.20 & 94.62 & 0.11 & 03.67 & 96.21 \\ \hline
val   & 0.41 & 03.77 & 95.82 & 0.01 & 04.01 & 95.98 \\ \hline
test  & 0.40 & 03.69 & 95.91 & 0.05 & 02.13 & 97.82 \\ \hline
\end{tabular}}
\caption{Comparing the number of hands present per frame in train, test, and val splits of EK100 and EGTEA.}
\label{tab:hands}
\end{table}

\section{Changing number of objects per frame}

We compare the performance of varying the number of objects per frame $N$ when training InAViT.
For EK100 (\cref{ekobj}), the performance improves when we increase the number of objects per frame to 4 but then deteriorates when we further increase it to 5. 
So, we use $N$=4 in all our experiments on EK100 in the main paper.
Similarly, for EGTEA (\cref{egteaobj}), we find that InAViT performs the best when we set $N$=2 objects per frame.

\begin{table}[]
\centering
\scriptsize
\begin{tabular}{c|c|c|c} 
\hline
\begin{tabular}[c]{@{}l@{}}\# Objects\\~per frame\end{tabular} & VERB & NOUN & ACTION\\
\hline
1 & 37.62 & 41.56 & 20.67\\ 
\hline
2 & 38.54 & 42.44 & 21.89\\ 
\hline
3 & 40.14 & 43.66 & 23.65\\ 
\hline
4 & \textbf{40.68 }& \textbf{44.13} & \textbf{24.65}\\
\hline
5 & 40.25 & 44.04 & 24.45 \\
\hline
\end{tabular}
\caption{Effect of changing the number of objects per frame $N$ on anticipation performance when training InAViT. Metric is Mean Recall@5 evaluated on EK100 validation set.}
\label{ekobj}
\end{table}

\begin{table}[]
\centering
\begin{tabular}{c|c|c|c} 
\hline
\begin{tabular}[c]{@{}l@{}}\# Objects\\~per frame\end{tabular} & VERB & NOUN & ACTION \\ 
\hline
1 & 78.9 & 75.8 & 65.7 \\ \hline
2 & \textbf{79.3} & \textbf{77.6} & \textbf{67.8} \\ \hline
3 & 78.5 & 76.0 & 64.8 \\ \hline
\end{tabular}
\caption{Changing number of objects per frame $N$ on EGTEA dataset. Metric is Top-1 Accuracy.}
\label{egteaobj}
\end{table}

\begin{table}
\resizebox{\linewidth}{!}{
\begin{tabular}{lcccccc} 
\hline
\multirow{2}{*}{Method} & \multicolumn{3}{c}{Top-1 Accuracy (\%)} & \multicolumn{3}{c}{Mean Class Accuracy (\%)} \\ 
\cline{2-7}
 & VERB & NOUN & ACT. & VERB & NOUN & ACT. \\ 
\hline
I3D-Res50~\cite{carreira2017quo} & 48.0 & 42.1 & 34.8 & 31.3 & 30.0 & 23.2~ \\ 
\hline
FHOI~\cite{liu2020forecasting} & 49.0 & 45.5 & 36.6 & 32.5 & 32.7 & 25.3~ \\ 
\hline
RU-LSTM~\cite{furnari2020rolling} & 50.3 & 48.1 & 38.6 & - & - & - \\ \hline
AFFT \cite{zhong2023anticipative} & 53.4 & 50.4 & 42.5 & 42.4 & 44.5 & 35.2  \\ 
\hline
AVT~\cite{girdhar2021anticipative} & 54.9 & 52.2 & 43.0 & 49.9 & 48.3 & 35.2 \\ \hline
Abs. Goal~\cite{roy2022predicting} & 64.8 & 65.3 & 49.8 & 63.4 & 55.6 & 37.4\\
\hline
MF$^{*}$ & 77.8 & 75.6 & 66.6 & 77.5 & 72.1 & 56.9 \\ \hline
ORVIT-MF$^{*}$ & 78.8 & 76.3 & 67.3 & 78.8 & 75.8 & 57.2  \\ \hline
\hline
InAViT (Ours) & \textbf{79.3} & \textbf{77.6} & \textbf{67.8} & \textbf{79.2} & \textbf{76.9} & \textbf{58.2} \\ \hline
\end{tabular}
}
\caption{Comparison of anticipation performance on EGTEA Gaze+.}
\label{tab:sota_egtea_sup}
\end{table}

\section{More attention qualitative results}
\begin{figure}
    \centering
    \begin{tabular}{c}
    \includegraphics[width=0.97\linewidth]{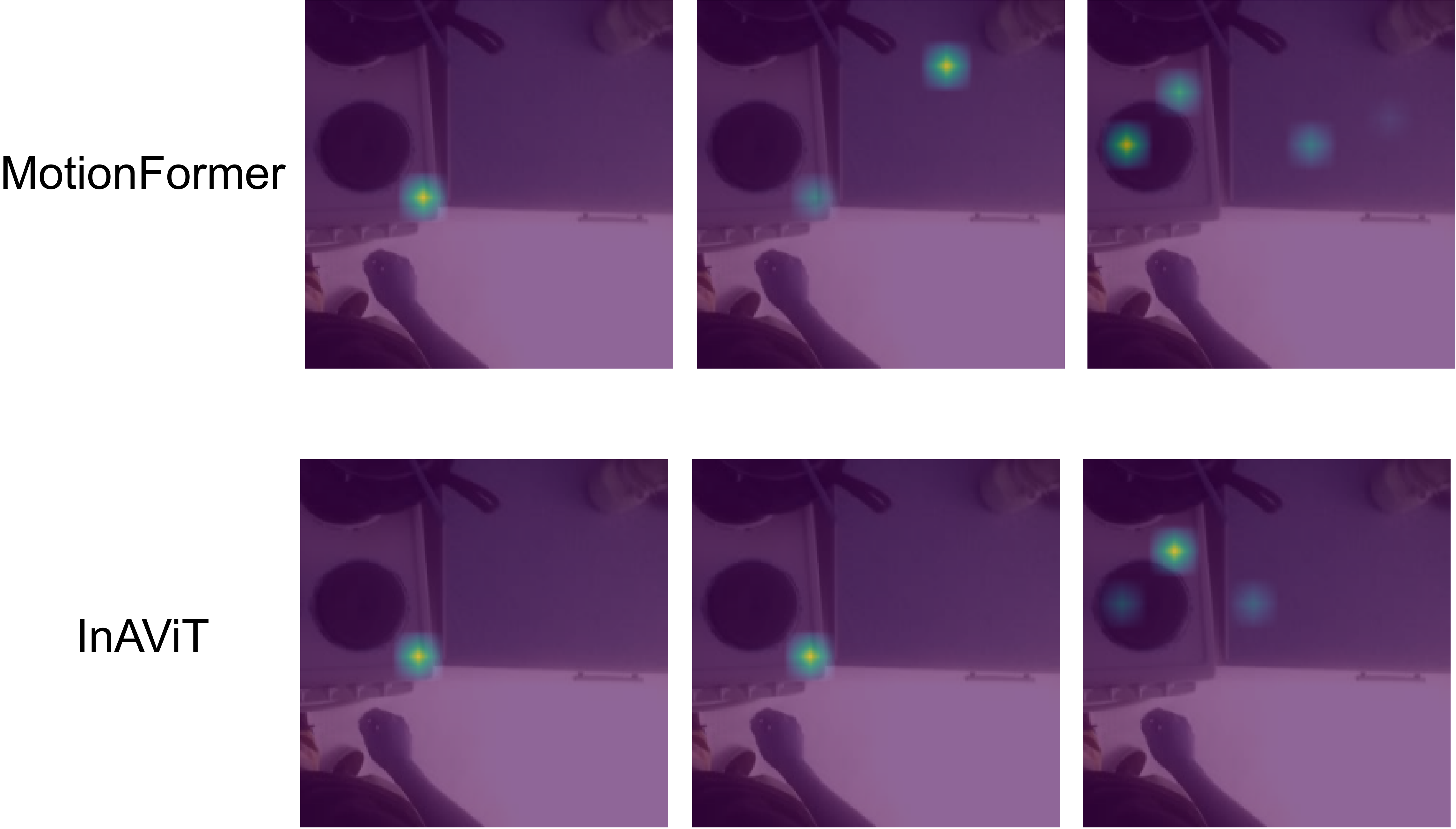} \\
     (a) Next action: \textit{turn knob} \\
    \includegraphics[width=0.97\linewidth]{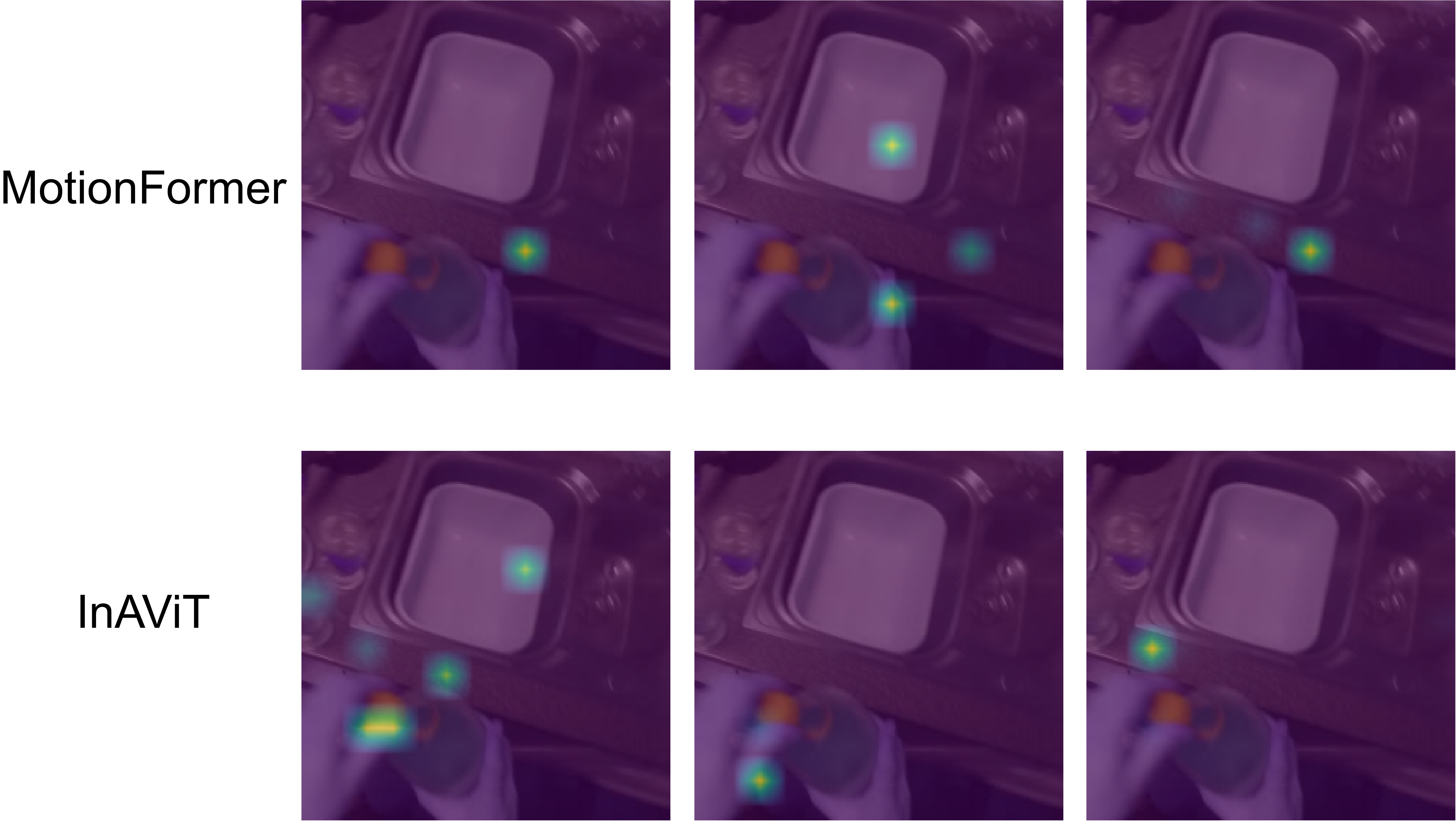} \\
    (b) Next action: \textit{open bottle}
    \end{tabular}
    \caption{(a) InAViT focuses near the knob when predicting the next action \textit{turn knob}. (b) InAViT focuses on the bottle cap when predicting the next action of \textit{open bottle}.}
    \label{morequal}
\end{figure}

In \cref{morequal}, we show more qualitative results of attention outputs  comparing MotionFormer and InAViT on two more actions - \textit{turn knob} and \textit{open bottle}. 
InAViT is able to focus on the important regions relevant for the next action in both cases better than MotionFormer.


\section{Full ablation table}
In \cref{supp:ablation}, we present the full ablation table as an addendum to Tab. 1 in the main paper. 
We include the verb and noun anticipation performance and observe similar trends as action anticipation. 
SCA+CI+ICV performs the best on both verb and noun anticipation.

\begin{table}[!ht]
\centering
\scriptsize
\begin{tabular}{lrrrrrrrrr}
\hline
\multirow{2}{*}{Method} & \multicolumn{3}{c}{Overall (\%)} & \multicolumn{3}{c}{Unseen (\%)} & \multicolumn{3}{c}{Tail (\%)} \\ 
\cline{2-10}
 & Verb & Noun & Action & Verb & Noun & Action & Verb & Noun & Action \\ 
\hline
SCA & 28.06 & 28.01 & 12.66 & 28.73 & 34.51 & 15.49 & 19.14 & 13.79 & 6.03 \\
\hline
SCA + CI & 37.63 & 38.71 & 14.21 & 34.93 & 38.89 & 14.26 & 30.68 & 29.10 & 9.12 \\
\hline
SCA+ICV & 48.14 & 47.71 & 22.21 & 43.61 & 46.44 & 20.85 & 42.49 & 37.83 & 17.07 \\
\hline

SCA + CI + TA & \textbf{49.14} & \textbf{49.97} & \textbf{23.75} & \textbf{44.36} & \textbf{49.28} & \textbf{23.49} & \textbf{43.17} & \textbf{39.91} & \textbf{18.11} \\
\hline
\multicolumn{10}{c}{(a) Component-wise validation of InAViT} \\
\hline
UB+CI+ICV & 45.16 & 47.99 & 22.75 & 42.9 & 48.05 & 22.14 & 38.54 & 36.94 & 17.04 \\
\hline
SOT+CI+ICV & 44.39 & 47.44 & 22.48 & 42.74 & 47.07 & 20.56 & 37.83 & 36.86 & 17.46 \\
\hline
\multicolumn{10}{c}{(b) Comparing interaction modeling methods} \\ 
\hline
SCA-(Hand)+CI+ICV & 47.44 & 48.91 & 23.27 & 43.42 & 48.07 & 23.21 & 41.08 & 38.27 & 17.57 \\
\hline
SCA-(Obj)+CI+ICV & 46.03 & 47.75 & 22.49 & 42.44 & 47.44 & 22.23 & 39.76 & 37.04 & 16.73 \\
\hline
\multicolumn{10}{c}{(c) Comparing refined hand vs. object as interaction tokens} \\ 
\hline
SCA+CI(Mask FG)+ICV & 29.67 & 25.38 & 8.05 & 24.02 & 23.19 & 5.92 & 24.01 & 16.57 & 5.84 \\
\hline
SCA+ Concat +ICV & 46.32 & 48.71 & 22.14 & 42.47 & 49.01 & 23.47 & 39.78 & 37.91 & 17.24 \\
\hline
\multicolumn{10}{c}{(d)  Effect of context infusion} \\
\end{tabular}
\caption{Full ablation of InAViT including verb and noun results on Action anticipation on EK100 evaluation server. 
CI=Context infusion, CI(Mask FG) = Context Infusion with foreground (hands and objects) masked out, Concat = Context infusion by concatenating context tokens with interaction tokens}
\label{supp:ablation}
\end{table}